\documentclass[letterpaper]{article}
\usepackage[preprint]{nips_2018}

\usepackage{booktabs} 
\usepackage{amssymb} 
\usepackage{amsmath} 
\usepackage{bm} 
\usepackage{threeparttable} 
\usepackage{url}
\usepackage{color}
\usepackage{graphicx}
\usepackage{subfigure}

\title{Modelling the Dynamic Joint Policy of Teammates \\ with Attention Multi-agent DDPG}

\author{
  Hangyu Mao \\
  Peking University \\
  \texttt{hy.mao@pku.edu.cn} \\
  \And
  Zhengchao Zhang \\
  Peking University \\
  \texttt{zhengchaozhang@pku.edu.cn} \\
  \And
  Zhen Xiao\thanks{The contact author.} \\
  Peking University \\
  \texttt{xiaozhen@net.pku.edu.cn} \\
  \And
  Zhibo Gong \\
  Huawei Technologies Co., Ltd. \\
  \texttt{gongzhibo@huawei.com} \\
}

\begin{document}


\maketitle

\begin{abstract}
Modelling and exploiting teammates' policies in cooperative multi-agent systems have long been an interest and also a big challenge for the reinforcement learning (RL) community. The interest lies in the fact that if the agent knows the teammates' policies, it can adjust its own policy accordingly to arrive at proper cooperations; while the challenge is that the agents' policies are changing continuously due to they are learning concurrently, which imposes difficulty to model the dynamic policies of teammates accurately. In this paper, we present \emph{ATTention Multi-Agent Deep Deterministic Policy Gradient} (ATT-MADDPG) to address this challenge. ATT-MADDPG extends DDPG, a single-agent actor-critic RL method, with two special designs. First, in order to model the teammates' policies, the agent should get access to the observations and actions of teammates. ATT-MADDPG adopts a centralized critic to collect such information. Second, to model the teammates' policies using the collected information in an effective way, ATT-MADDPG enhances the centralized critic with an attention mechanism. This attention mechanism introduces a special structure to explicitly model the dynamic joint policy of teammates, making sure that the collected information can be processed efficiently. We evaluate ATT-MADDPG on both benchmark tasks and the real-world packet routing tasks. Experimental results show that it not only outperforms the state-of-the-art RL-based methods and rule-based methods by a large margin, but also achieves better performance in terms of scalability and robustness.
\end{abstract}

\section{Introduction}
There are many real-world tasks involving multiple agents, such as the network packet routing \citep{vicisano1998tcp,tao2001multi}, the autonomous intersection management \citep{dresner2008multiagent} and the Poker games \citep{billings1998opponent}. In the past decades, researchers have made continuous attempts to apply reinforcement learning (RL) \citep{sutton1998introduction} to deal with these multi-agent tasks, because solving these tasks using a learning-based method is a crucial step to build artificial intelligent systems. Nevertheless, it remains an open question due to many challenges, for example, the partial observability of agents, the cooperation and competition among agents, the changing number of agents, and etc.

In this paper, we focus on the cooperative distributed multi-agent RL setting. In cooperative setting, the agents need to take collaborative actions to achieve a shared goal. In distributed setting, the agents are located in different areas with partial observability. A representative task is the packet routing where the routers are treated as the autonomous agents and the goal is to transmit the packets using as less resource as possible.

Even in this simplified setting, it is still difficult to handle such tasks due to the complex agent modelling problem \citep{albrecht2018autonomous}. Specifically, if the agent maintains the models about teammates' policies, it can adjust its own policy accordingly to achieve a proper cooperation. However, since the agents are learning concurrently in the same environment, their policies are changing continuously. This kind of dynamically changing policy is very hard to model in an accurate manner. Even if one can manage to do it, it is easily outdated anyway.

In fact, modelling and exploiting teammates' policies have long been an interest for the RL community, as summarized in this excellent survey \citep{albrecht2018autonomous}. Nevertheless, most methods are introduced in the Game Theory \citep{ganzfried2011game} or simple grid-world settings, and they usually model each teammates' policy separately. It is hard to scale these methods to real-world applications like the network packet routing.

Recently, Deep Reinforcement Learning (DRL) has been explored for large scale tasks. In order to achieve generalization in tasks with large state space and action space, DRL-based methods adopt \emph{deep neural network} as function approximator to generate similar actions for similar states. However, the existing DRL-based agent modelling methods \citep{cite_DRON,cite_LOLA,cite_SOM,cite_MeanField,cite_DPIQN} mostly focus on improving the deep Q-network (DQN) \citep{DQN}, and they usually learn centralized policies. To apply the centralized policy in distributed systems, the agents have to exchange information during execution, which is too costly or even unattainable in many cases \citep{roth2005reasoning,Zhang2013Coordinating,chen2017decentralized,dobbe2017fully}. In addition, the DQN-based methods target at addressing tasks with discrete action space. Other DQN-based methods \citep{VDN,QMIX} or researches \citep{foerster2017counterfactual,lowe2017multi,chu2017parameter} based on actor-critic RL algorithm can generate decentralized policies, but they do not explicitly build models for other agents. Instead, they investigate other topics such as the credit assignment among multiple agents.

In this paper, we present \emph{ATTention Multi-Agent Deep Deterministic Policy Gradient} (ATT-MADDPG) to address the complex agent modelling problem. In contrast to previous works, ATT-MADDPG \emph{explicitly} model the dynamic \emph{joint policy} of teammates in an \emph{adaptive manner}, and it is designed for training \emph{decentralized policies} to handle large-scale distributed tasks with \emph{continuous action space}.

Specifically, ATT-MADDPG extends DDPG \citep{DDPG}, a single-agent actor-critic RL algorithm, with two special designs. First, as a necessary step to do agent modelling, the agent should get access to the observations and actions of teammates. ATT-MADDPG adopts a \emph{centralized critic} to collect these information. Second, in order to make sure that the collected information can be processed in an effective way to model the teammates' policies, ATT-MADDPG further embeds an \emph{attention mechanism} into the centralized critic. This attention mechanism introduces a special structure to \emph{explicitly} model the dynamic \emph{joint policy} of teammates in an \emph{adaptive manner}. Once the teammates change their policies, the associated attention weight will change adaptively, and the agent will adjust its policy quickly. Consequently, all agents will cooperate efficiently. In addition, since DDPG targets at continuous action space tasks, ATT-MADDPG can naturally deal with such tasks. Moreover, the policy is decentralized because we do not change the actor part of DDPG, and the actor can generate action based on its own observation history.

We evaluate ATT-MADDPG on the real-world packet routing tasks as well as benchmark cooperative navigation and predator prey tasks. In all tasks, ATT-MADDPG can obtain more rewards than both the state-of-the-art RL-based methods and rule-based methods. Experiments also show that ATT-MADDPG achieves better scalability and robustness. Furthermore, we conduct experiments on packet routing task to reveal some insights about the attention mechanism, and on cooperative navigation task to show the cooperation among the agents' policies.

Our main contributions can be summarized as follows.
\begin{itemize}
\item In contrast to most agent modelling methods, ATT-MADDPG trains a \emph{decentralized policy} for each agent to handle distributed tasks with \emph{continuous action}.
\item The proposed \emph{attention mechanism} introduces a special structure to \emph{explicitly} model the dynamic \emph{joint policy} of teammates in an \emph{adaptive manner}. To our knowledge, we are the first to do agent modelling in this novel way.
\item We empirically test ATT-MADDPG on both real-world tasks and benchmark tasks to show that it achieves good performance in terms of the reward, scalability and robustness.
\end{itemize}

\section{Background} \label{Background}
\textbf{DEC-POMDP.} We consider a multi-agent setting that can be formulated as DEC-POMDP \citep{bernstein2002complexity}. It is formally defined as a tuple $\langle N,S,\vec{A},T,\vec{R},\vec{O},Z,\gamma \rangle$, where $N$ is the number of agents; $S$ is the set of state $s$; $\vec{A}=[A_1, ..., A_N]$ represents the set of \emph{joint action} $\vec{a}$, and $A_i$ is the set of \emph{local action} $a_i$ that agent $i$ can take; $T(s'|s,\vec{a}): S \times \vec{A} \times S \rightarrow [0,1]$ represents the state transition function; $\vec{R}=[R_1, ..., R_N]: S \times \vec{A} \rightarrow \mathbb{R}^N$ is the \emph{joint reward} function; $\vec{O} = [O_1, ..., O_N]$ is the set of \emph{joint observation} $\vec{o}$ controlled by the observation function $Z: S \times \vec{A} \rightarrow \vec{O}$; $\gamma \in [0,1]$ is the \emph{discount factor}.

In a given state $s$, each agent takes an action $a_i$ based on its own observation (history) $o_i$, resulting in a new state $s'$ and a reward $r_i$ \footnote{In practice, we map observation history instead of current observation to an action. In cooperative setting, $r_i=r_j$ for different $i$ and $j$.}. The agent tries to learn a policy $\pi_{i}: O_i \times A_i \rightarrow [0,1]$ that can maximize $\mathbb{E}[G_i]$ where $G_i$ is the \emph{discount return} defined as $G_i = \sum_{t=0}^{H} \gamma^{t} r_{i}^{t}$, and $H$ is the time horizon. In addition, \textbf{we also assume that the environment is joint fully observable} \citep{bernstein2002complexity}, i.e., $s \triangleq \vec{o} = \langle o_{i}, \vec{o}_{-i} \rangle$ where $\vec{o}_{-i}$ is the joint observation (history) of teammates of agent $i$.

\textbf{Reinforcement Learning (RL).} RL \citep{sutton1998introduction} is generally used to solve special DEC-POMDP problems where $N=1$. In practice, the Q-value function $Q^{\pi}(s,a)$ is defined as
\begin{eqnarray}
    Q^{\pi}(s,a) &=& \mathbb{E}_{\pi}[G|S=s,A=a] \label{equ:Qfunction}
\end{eqnarray}
then the optimal policy is derived by $\pi^{*} = \arg\max_{\pi} Q^{\pi}(s,a)$.

Policy Gradient methods \citep{sutton2000policy} directly learn the parameterized policy $\pi_{\theta} = \pi(a|s;\theta)$, which is an approximation of any policy $\pi$. To maximize the objective $J(\theta) = \mathbb{E}_{s \sim p^{\pi}, a \sim \pi_{\theta}}[G]$, the parameters $\theta$ are adjusted in the direction of $\nabla_{\theta}J(\theta) = \mathbb{E}_{s \sim p^{\pi}, a \sim \pi_{\theta}}[\nabla_{\theta} \log\pi(a|s;\theta)Q^{\pi}(s,a)]$, where $p^{\pi}$ is the stable state distribution. We can use deep neural network $Q(s,a;w)$ to approximate $Q^{\pi}(s,a)$, resulting in the actor-critic algorithms \citep{konda2003onactor,grondman2012survey}. Both the parameterized actor $\pi(a|s;\theta)$ and critic $Q(s,a;w)$ are used during training, while \textbf{only the actor $\bm{\pi(a|s;\theta)}$ is needed during execution}. This merit will be used to train decentralized policies in our method.

Deterministic Policy Gradient (DPG) \citep{DPG} is a special actor-critic algorithm where the actor adopts a deterministic policy $\mu_{\theta}: S \rightarrow A$ and the action space $A$ is continuous. Deep DPG (DDPG) \citep{DDPG} uses deep neural networks to approximate $\mu_{\theta}(s)$ and $Q(s,a;w)$. DDPG is an off-policy method, which applies the \emph{target network} and \emph{experience replay} to stabilize training and to improve data efficiency. Specifically, the critic and actor are updated based on the following equations:
\begin{eqnarray}
    \delta =& \hspace{-0.5em} r + \gamma Q(s',a';w^{-})|_{a'=\mu_{\theta^{-}}(s')} - Q(s,a;w) \label{equ:DPG1} \\
    L(w) =& \hspace{-10.6em} \mathbb{E}_{(s,a,r,s') \sim D}[(\delta)^{2}] \label{equ:DPG2} \\
    \nabla_{\theta}J(\theta) =& \hspace{-1.8em} \mathbb{E}_{s \sim D}[\nabla_{\theta}\mu_\theta(s) * \nabla_{a}Q(s,a;w)|_{a=\mu_{\theta}(s)}] \label{equ:DPG3}
\end{eqnarray}
where $D$ is the replay buffer containing recent experience tuples $(s,a,r,s')$; $Q(s,a;w^{-})$ and $\mu_{\theta^{-}}(s)$ are the target networks whose parameters $w^{-}$ and $\theta^{-}$ are periodically updated by copying $w$ and $\theta$.

\begin{figure}[!htb]
    \centering
    \includegraphics[width=6.6cm]{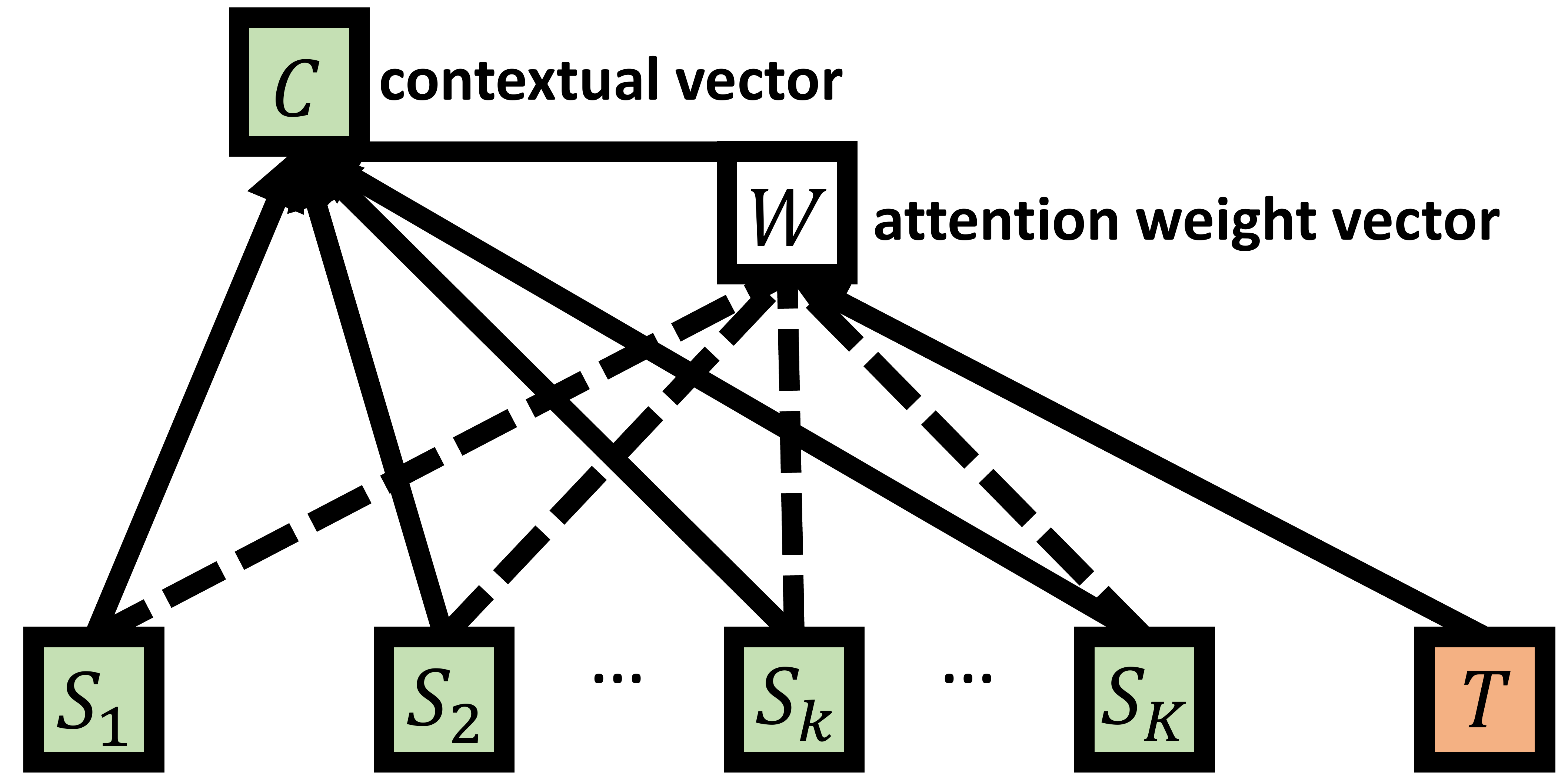}
    \caption{The Soft Attention \citep{cite_SoftAttention,cite_GlobalAttention}.}
    \label{fig:Attention}
\end{figure}
\textbf{Attention Mechanism.} The Soft Attention \citep{cite_SoftAttention} (sometimes referred as Global Attention \citep{cite_GlobalAttention}) is the most popular one as shown in Figure \ref{fig:Attention}. The inputs are several source vectors $[S_1,S_2,..,S_k,..,S_K]$ and one target vector $T$. \emph{The model can \textbf{adaptively} attend to more important $S_k$}, where the importance is measured by a user-defined function $f(T,S_k)$; and the important information contained in $S_k$ can be encoded into a contextual vector $C$ adaptively according to the normalized importance score $w_k$ as follows:
\begin{eqnarray}
    w_k = \frac{\exp(f(T,S_k))}{\sum_{i=1}^{K} \exp(f(T,S_i))}
    \hspace{0.6em} \text{;} \hspace{0.6em}
    C = \sum_{k=1}^{K} w_k S_k
\end{eqnarray}
Besides, the attention weight vector $W \triangleq [w_1,w_2,..,w_k,..,w_K]$ can also be seen as a \textbf{probability distribution} because $\sum_{k=1}^{K} w_k \equiv 1$. The ingenuity for generating a probability distribution adaptively will be applied in our method.

\section{Attention Multi-agent DDPG}\label{ATT-MADDPG}
Before digging into the details, we list the key variables used in this paper in Table \ref{tab:variables}. Please notice the differences between $\bm{\vec{\pi}_{-i}}$, $\bm{\vec{\pi}_{-i}(\vec{a}_{-i}|s)}$ and $\bm{\vec{\pi}_{-i}(\vec{A}_{-i}|s)}$.

\begin{table}[!htb]
    \newcommand{\tabincell}[2]{\begin{tabular}{@{}#1@{}}#2\end{tabular}}
    \centering
    \caption{The key variables used in this paper.}
    \label{tab:variables}
    \begin{threeparttable}
        \begin{tabular}{|l|l|}
            \hline
            \bf $\vec{a}$ & The joint action of all agents. \\
            \hline
            \bf $a_i$ & The local action of agent $i$. \\
            \hline
            \bf $\vec{a}_{-i}$ & The joint action of teammates of agent $i$. \\
            \hline
            \multicolumn{2}{|l|}{The action set $\vec{A}$, $A_i$, $\vec{A}_{-i}$ are denoted similarly.}  \\
            \hline
            \multicolumn{2}{|l|}{The observation (history) $\vec{o}$, $o_i$, $\vec{o}_{-i}$ are denoted similarly.}  \\
            \hline
            \multicolumn{2}{|l|}{The policy $\vec{\pi}$, $\pi_i$, $\vec{\pi}_{-i}$ are denoted similarly.}  \\
            \hline
            \hline
            \bf $s'$ & The next state after $s$. \\
            \hline
            \multicolumn{2}{|l|}{$\vec{o}'$, $o_i'$, $\vec{o}_{-i}'$, $\vec{a}'$, $a_i'$, and $\vec{a}_{-i}'$ are denoted similarly.}  \\
            \hline
            \hline
            \bf $\bm{\vec{\pi}_{-i}}$ & The \textbf{joint policy} of teammates of agent $i$. \\
            \hline
            \bf $\bm{\vec{\pi}_{-i}(\vec{a}_{-i}|s)}$ & \tabincell{l}{The \textbf{probability value} for generating $\vec{a}_{-i}$ \\ under policy $\vec{\pi}_{-i}$. $\Sigma_{\vec{a}_{-i} \in \vec{A}_{-i} }\vec{\pi}_{-i}(\vec{a}_{-i}|s) = 1$.} \\
            \hline
            \bf $\bm{\vec{\pi}_{-i}(\vec{A}_{-i}|s)}$ & \tabincell{l}{The \textbf{probability distribution} over the  \\ joint action space $\vec{A}_{-i}$ under policy $\vec{\pi}_{-i}$.} \\
            \hline
        \end{tabular}
    \end{threeparttable}
\end{table}

\subsection{The Overall Approach} \label{MADDPG}
\begin{figure}[!htb]
    \centering
    \includegraphics[width=8.6cm]{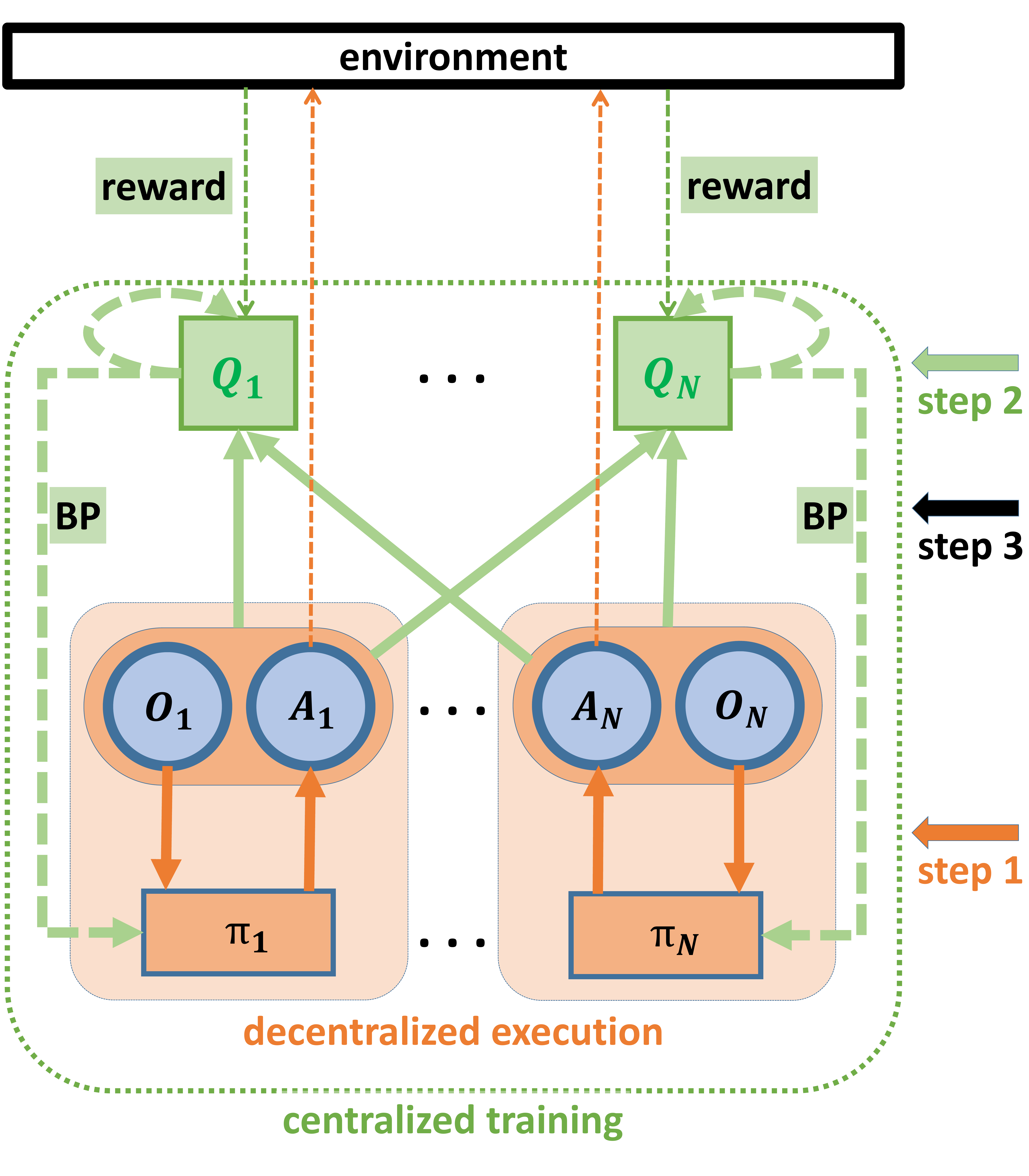}
    \caption{The overall approach of ATT-MADDPG.}
    \label{fig:ATT_MADDPG_Overall}
\end{figure}

The proposed ATT-MADDPG extends the actor-critic RL algorithm with a centralized critic and an attention mechanism. To make our method more easy to understand, we present the overall approach without considering the attention mechanism. We will introduce it in the next section.

Specifically, as can be seen from Figure \ref{fig:ATT_MADDPG_Overall}, the centralized critic $Q_i$ (i.e., the Q-value function that is related to agent $i$) can get access to the observations and actions of \emph{all} agents, while the independent actor $\pi_i$ can only get access to its own observation $o_i$. Accordingly, ATT-MADDPG works as follows during training.

Step 1: the actors $\pi_i$ generate the actions $a_i$ based on their \emph{own} observations $o_i$ to interact with the environment.

Step 2: the centralized critics estimate the Q-values $Q_i$ based on the observations and actions of \emph{all} agents.

Step 3: after receiving the feedback reward from the environment, the actors and critics are jointly trained using back propagation (BP) based on Equation \ref{equ:ATT-MADDPG1}, \ref{equ:ATT-MADDPG2}, and \ref{equ:ATT-MADDPG3}.

Although the overall approach is simple, it has great ability to address the agent modelling problem in distributed setting: (1) note that only step 1 is needed during execution, thus the independent actor $\pi_i$ can learn decentralized policies that are suitable for the distributed setting; (2) generally, there is no way to model the policies of other agents without accessing their observations $\vec{o}_{-i}$ and actions $\vec{a}_{-i}$; in step 2, the centralized critic $Q_i$ is designed to collect $\vec{o}_{-i}$ and $\vec{a}_{-i}$, which forms the necessary foundation to do agent modelling. Moreover, with centralized critics, the agents can be trained with stable reward signal $r_{i}$, hence our method can also relieve the non-stationary problem \citep{weinberg2004best,hernandez2017survey}\footnote{The reason is that a joint action $\vec{a} = \langle a_{i}, \vec{a}_{-i} \rangle$ taken in a given state $s \triangleq \vec{o} = \langle o_{i}, \vec{o}_{-i} \rangle$ can invariably result in the same $r_{i}$ and $s'$ with deterministic probability, which is regardless of the changing policies of other teammates. More discussion can be found in \citep{lowe2017multi,foerster2017counterfactual,mao2017accnet,chu2017parameter,gupta2017cooperative}, which also adopt centralized critics (but do not study agent modelling).}. 

\subsection{The Attention Critic}
\begin{figure}[!htb]
    \centering
    \includegraphics[width=12.6cm]{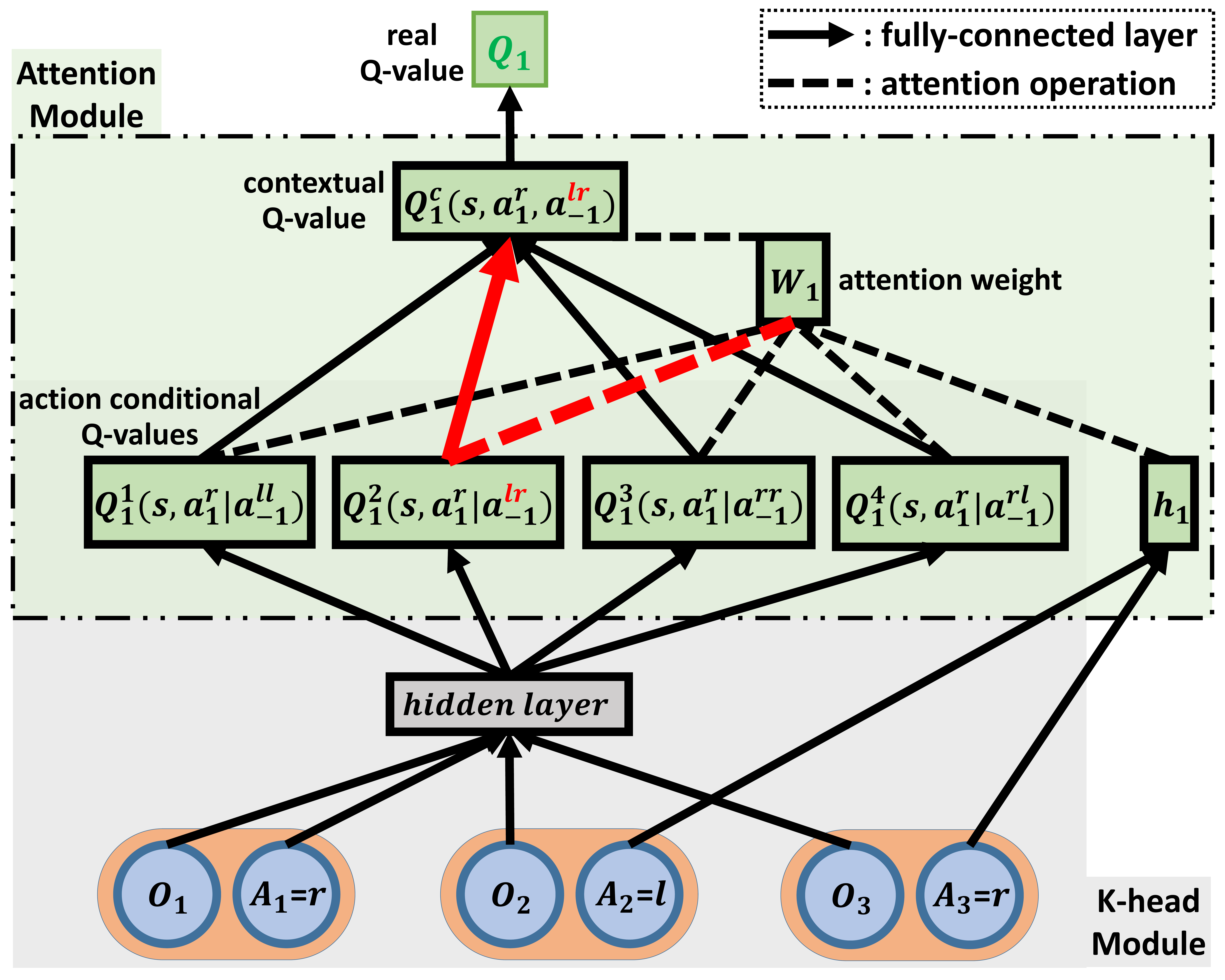}
    \caption{The attention critic of ATT-MADDPG. We show the detailed generation of $Q_1$ using a three-agent example: the discrete action space is $\{l, r\}$, and the agents prefer to take the actions $r$, $\color{red}{l}$, and $\color{red}{r}$, respectively. In this case, the second \emph{action conditional Q-value} $Q^{2}_{1}$ will contribute more weights to the computation of the \emph{contextual Q-value} $Q^{c}_{1}$, as indicated by thicker red links. \emph{\textbf{Note that}} we call $Q_i$ the \emph{real Q-value}, $Q^{c}_{i}$ the \emph{contextual Q-value}, and $Q^{k}_{i}$ the \emph{action conditional Q-value}. The difference is that $Q^{c}_{i}$ and $Q^{k}_{i}$ are multi-dimensional vectors, while $Q_i$ is the real scalar Q-value used in Equation \ref{equ:ATT-MADDPG1}, \ref{equ:ATT-MADDPG2} and \ref{equ:ATT-MADDPG3}.}
    \label{fig:QvalueGeneration}
\end{figure}

To arrive at a proper cooperation, the agent is expected to model the teammates' policies and to adjust its own policy accordingly. We design and embed a kind of Soft Attention into the centralized critic, making sure that the dynamic joint policies of teammates can be modelled adaptively.

To make our design more easy to understand, we introduce it based on the assumption that the action is discrete. The extension to continuous action is presented in Section \ref{Implementation}.

Recall that the environment is influenced by $\vec{a}$ in multi-agent setting. From the perspective of agent $i$, the outcome of $a_{i}$ taken in $s$ is dependent on $\vec{a}_{-i}$. Therefore, similar to the definition of $Q^{\pi}(s,a)$ in Equation \ref{equ:Qfunction}, we define the \emph{Q-value function \textbf{relative to} the joint policy of teammates} as $Q_{i}^{\pi_i|\vec{\pi}_{-i}}(s,a_i)$ as previous studies \citep{cite_DRON,Banerjee2007Reaching}, and our new objective is to find the optimal policy $\pi_i^{*} = \arg\max_{\pi_i} Q_{i}^{\pi_i|\vec{\pi}_{-i}}(s,a_i)$. Mathematically, $Q_{i}^{\pi_i|\vec{\pi}_{-i}}(s,a_i)$ can be calculated by\footnote{The detailed derivation can be found in \citep{cite_DRON} and the Appendix.}
\begin{eqnarray}
    Q_{i}^{\pi_i|\vec{\pi}_{-i}}(s,a_i) =& \hspace{-4.6em} \mathbb{E}_{\vec{a}_{-i} \sim \vec{\pi}_{-i}}[Q_{i}^{\pi_i}(s,a_i,\vec{a}_{-i})] \label{eqnarray:attentionQ1}\\
    =& \Sigma_{\vec{a}_{-i} \in \vec{A}_{-i}}[\vec{\pi}_{-i}(\vec{a}_{-i}|s) Q_{i}^{\pi_i}(s,a_i,\vec{a}_{-i})] \label{eqnarray:attentionQ2}
\end{eqnarray}

Equation \ref{eqnarray:attentionQ2} implies that in order to estimate $Q_{i}^{\pi_i|\vec{\pi}_{-i}}(s,a_i)$, the critic network of agent $i$ should have the abilities:

(1) to estimate $Q_{i}^{\pi_i}(s,a_i,\vec{a}_{-i})$ for each $\vec{a}_{-i} \in \vec{A}_{-i}$.

(2) to calculate the expectation of all $Q_{i}^{\pi_i}(s,a_i,\vec{a}_{-i})$\footnote{The expectation is equivalent to the weighted summation, and the weight of $Q_{i}^{\pi_i}(s,a_i,\vec{a}_{-i})$ is $\vec{\pi}_{-i}(\vec{a}_{-i}|s)$ as shown in Equation \ref{eqnarray:attentionQ2}.}.

To estimate $Q_{i}^{\pi_i}(s,a_i,\vec{a}_{-i})$ for each $\vec{a}_{-i} \in \vec{A}_{-i}$, we design a \textbf{\emph{$\bm{K}$-head Module}} where $K$=$|\vec{A}_{-i}|$. As shown at the bottom of Figure \ref{fig:QvalueGeneration}, the $K$-head Module generates $K$ \emph{action conditional Q-value} $Q_{i}^{k}(s,a_i|\vec{a}_{-i};w_i)$ for each $\vec{a}_{-i}$ to approximate the true $Q_{i}^{\pi_i}(s,a_i,\vec{a}_{-i})$, where $w_i$ is the parameters of the critic network of agent $i$. Specifically, $Q_{i}^{k}(s,a_i|\vec{a}_{-i};w_i)$ is generated using $a_i$ and all observations $\langle o_{i}, \vec{o}_{-i} \rangle = \vec{o} \triangleq s$; as for the information about $\vec{a}_{-i}$, it is provided by an additional hidden vector $h_i(w_i)$, which will be introduced shortly\footnote{This is why we use $Q_{i}^{k}(s,a_i|\vec{a}_{-i};w_i)$ instead of $Q_{i}^{k}(s,a_i,\vec{a}_{-i};w_i)$ to represent the defined \emph{action conditional Q-value}.}.

To calculate the expectation of all $Q_{i}^{\pi_i}(s,a_i,\vec{a}_{-i})$, the weights $\vec{\pi}_{-i}(\vec{a}_{-i}|s)$ of all $Q_{i}^{\pi_i}(s,a_i,\vec{a}_{-i})$ are also required as indicated by Equation \ref{eqnarray:attentionQ2}. However, it is hard to approximate these weights. On one hand, for different $s$, the teammates will take different $\vec{a}_{-i}$ with different probabilities $\vec{\pi}_{-i}(\vec{a}_{-i}|s)$ based on the policy $\vec{\pi}_{-i}$. On the other hand, the policy $\vec{\pi}_{-i}$ is changing continuously, because the agents are learning concurrently to adapt to each other.

We propose to approximate all $\vec{\pi}_{-i}(\vec{a}_{-i}|s) \in \vec{\pi}_{-i}(\vec{A}_{-i}|s)$ \emph{jointly} by a weight vector $W_i(w_i) \triangleq [W_i^{1}(w_i),...,W_i^{K}(w_i)]$, where $w_i$ is the parameters of the critic network of agent $i$. That is to say, we use $W_i(w_i)$ to approximate the \emph{probability distribution} $\vec{\pi}_{-i}(\vec{A}_{-i}|s)$, rather than approximating each \emph{probability value} $\vec{\pi}_{-i}(\vec{a}_{-i}|s)$ separately. A good $W_i(w_i)$ should satisfy the following conditions: (1) $\Sigma_{k=1}^{K} W_i^{k}(w_i) \equiv 1$, such that $W_i(w_i)$ is a probability distribution indeed; (2) $W_i(w_i)$ can change adaptively when the joint policy of teammates $\vec{\pi}_{-i}$ is changed, such that $W_i(w_i)$ can really model the teammates' joint policy in an adaptive manner.

Recall that the attention mechanism is intrinsically suitable for generating a probability distribution in an adaptive manner (please refer Section \ref{Background}), so we leverage it to design an \textbf{\emph{Attention Module}}. As shown at the middle of Figure \ref{fig:QvalueGeneration}, Attention Module works as follows.

Firstly, a hidden vector $h_i(w_i)$ is generated based on all actions of teammates (i.e., $\vec{a}_{-i}$).

Then, the attention weight vector $W_i(w_i)$ is generated by comparing $h_i(w_i)$ with all action conditional Q-values $Q_{i}^{k}(s,a_i|\vec{a}_{-i};w_i)$. Specifically, we apply the dot score function \citep{cite_GlobalAttention} to calculate the element $W_i^{k}(w_i) \in W_i(w_i)$:
\begin{eqnarray}
    W_i^{k}(w_i) = \frac{\exp(h_i(w_i) Q_i^{k}(s,a_i|\vec{a}_{-i};w_i))}{\sum_{k=1}^{K} \exp(h_i(w_i) Q_i^{k}(s,a_i|\vec{a}_{-i};w_i))} \label{equ:AttentionWeight}
\end{eqnarray}

Lastly, the \emph{contextual Q-value} $Q_i^{c}(s,a_i,\vec{a}_{-i};w_i)$ is calculated as a weighted summation of $W_i^{k}$ and $Q_i^{k}$:
\begin{eqnarray}
    Q_i^{c}(s,a_i,\vec{a}_{-i};w_i) = \sum_{k=1}^{K} W_i^{k}(w_i) Q_i^{k}(s,a_i|\vec{a}_{-i};w_i) \label{eqnarray:contextualQ}
\end{eqnarray}

\textbf{Summary:} Teammates have been considered in Equation \ref{eqnarray:attentionQ2}, while Equation \ref{eqnarray:contextualQ} is an approximation of Equation \ref{eqnarray:attentionQ2}, because $W_i^{k}(w_i)$ and $Q_i^{k}(s,a_i|\vec{a}_{-i};w_i)$ can learn to approximate $\vec{\pi}_{-i}(\vec{a}_{-i}|s)$ and $Q_i^{\pi_i}(s,a_i,\vec{a}_{-i})$, respectively. Thus, the agents controlled by ATT-MADDPG can cooperate efficiently. 

\subsection{Key Implementation} \label{Implementation}
\textbf{Attention Module.} After getting the contextual Q-value $Q_i^{c}(s,a_i,\vec{a}_{-i};w_i)$, we need to transform the multi-dimensional $Q_i^{c}$ into a scalar \emph{real Q-value} $Q_i$ using a fully-connected layer with one output neuron, as shown at the top of Figure \ref{fig:QvalueGeneration}.

The reason is that many researches have shown that multi-dimensional vector works better than scalar when implementing the Soft Attention \citep{cite_SoftAttention}. In our Attention Module, we also find that vector works better than scalar, so the $Q^{c}_{i}$, $Q^{k}_{i}$, $h_i(w_i)$ and $W_i(w_i)$ are all implemented using vectors. However, the standard RL adopts a scalar real Q-value $Q_i$, thus we should transform $Q_i^{c}$ into a scalar real Q-value $Q_i$.

\textbf{$\bm{K}$-head Module.} We have limited the above discussion to discrete action space. A natural question is that should we generate one $Q^{k}_{i}(s,a_i|\vec{a}_{-i};w_i)$ for each $\vec{a}_{-i} \in \vec{A}_{-i}$? What if the action space is continuous?

In fact, \emph{\textbf{there is no need to set $\bm{K=|\vec{A}_{-i}|}$}}. Many researchers have shown that only a small set of actions are crucial in most cases, and the conclusion is suitable for both continuous \citep{DPG} and discrete \citep{wang2015dueling} action space environments.

We argue that if $Q^{k}_{i}(s,a_i|\vec{a}_{-i};w_i)$ could group similar $\vec{a}_{-i}$ (i.e., representing different but similar $\vec{a}_{-i}$ using one Q-value head), it will be much more efficient. As deep neural network is an universal function approximator \citep{UniversalApproximator1,UniversalApproximator2,UniversalApproximator3}, we expect that our method can possess this ability. Further analysis in Section \ref{AttentionAnalysis} also indicates that our hypothesis is reasonable. Hence, we adopt a small $K$ (e.g., 4 or 8) even in tasks with continuous action space.

\textbf{Parameter Updating Method.} Since the critic network has considered the observations and actions of all agents, the network's output (i.e., the real Q-value $Q_{i}$) can be represented as $Q_{i}(\langle o_i,\vec{o}_{-i} \rangle,a_{i},\vec{a}_{-i};w_{i})$. Therefore, we can extend Equation \ref{equ:DPG1}, \ref{equ:DPG2} and \ref{equ:DPG3} into multi-agent formulations:
\begin{eqnarray}
    \delta_i =& \hspace{-0.5em} r_i+\gamma Q_i(\langle o'_i,\vec{o}'_{-i} \rangle,a'_i,\vec{a}'_{-i};w_i^{-})|_{a'_j=\mu_{\theta_j^{-}}(o'_j)} \nonumber \\
    & \hspace{-7.0em} - \hspace{0.3em} Q_{i}(\langle o_i,\vec{o}_{-i} \rangle,a_{i},\vec{a}_{-i};w_{i}) \label{equ:ATT-MADDPG1} \\
    L(w_i) =& \hspace{-4.5em} \mathbb{E}_{(o_i,\vec{o}_{-i},a_i,\vec{a}_{-i},r_i,o'_i,\vec{o}'_{-i}) \sim D}[(\delta_i)^{2}] \label{equ:ATT-MADDPG2} \\
    \nabla_{\theta_i}J(\theta_i) =& \hspace{-8.2em} \mathbb{E}_{(o_i,\vec{o}_{-i}) \sim D}[\nabla_{\theta_i}\mu_{\theta_i}(o_i) \nonumber \\
    & \hspace{-0.8em} * \hspace{0.2em} \nabla_{a_i}Q_{i}(\langle o_i,\vec{o}_{-i} \rangle,a_{i},\vec{a}_{-i};w_{i})|_{a_j=\mu_{\theta_j}(o_j)}] \label{equ:ATT-MADDPG3}
\end{eqnarray}

In practice, we adopt the \emph{centralized training with decentralized execution} paradigm \citep{oliehoek2008optimal,foerster2017counterfactual,lowe2017multi,chu2017parameter} to train and deploy our model, thus the information in the above equations can be collected easily. Besides, the $K$-head Module and Attention Module are submodules embedded in the centralized critic, so they can be optimized jointly with the agent's policy in an end-to-end manner using back propagation.

\subsection{The Discussion}
Our attention critic has great ability to \emph{explicitly} model the dynamic \emph{joint policy} of teammates in an \emph{adaptive manner}. This can be understood from three perspectives.

The first perspective is the \emph{joint policy}. Equation \ref{equ:AttentionWeight} makes sure that $\Sigma_{k=1}^{K} W_i^{k}(w_i) \equiv 1$, thus $W_i(w_i)$ must be able to represent the probability distribution $\vec{\pi}_{-i}(\vec{A}_{-i}|s)$ of a specific joint policy $\vec{\pi}_{-i}$.

The second perspective is the \emph{adaptive manner}. That is to say, $W_i(w_i)$ can react to the teammates' dynamic policies adaptively. The reason is that the action conditional Q-value $Q^{k}_{i}(s,a_i|\vec{a}_{-i};w_i)$ has considered all actions of the agent team, thus its values can be estimated using the experience tuple $(s, \langle a_i,\vec{a}_{-i} \rangle, r_i, s')$ $\triangleq$ $(\langle o_i,\vec{o}_{-i} \rangle, \langle a_i,\vec{a}_{-i} \rangle, r_i, \langle o'_i,\vec{o}'_{-i} \rangle)$, which is independent of the current $\vec{\pi}_{-i}$. It means that $Q^{k}_{i}$ has no need to \emph{shift} its values even if $\vec{\pi}_{-i}$ has changed (yet $Q^{k}_{i}$ still need to be learned). Given a stable $Q^{k}_{i}$, the attention weight $W_i(w_i)$ can adapt to different $\vec{\pi}_{-i}$ easily, and the agent will adjust its policy quickly.

The last perspective is that the critic network is designed based on mathematical analysis, which introduces a special structure to \emph{explicitly} approximate Equation \ref{eqnarray:attentionQ2}. This is similar to the renowned Dueling Network \citep{wang2015dueling}, which \emph{explicitly} approximates the Q-value as the summation of the advantage and the baseline (i.e., $Q(s,a)=A(s,a) + V(s)$). In contrast, if the centralized critic is implemented using fully-connected network like previous studies \citep{foerster2017counterfactual,mao2017accnet,lowe2017multi}, it will be hard for the fully-connected critic network to accomplish such meticulous task.

\section{Experiment}\label{Experiment}
The experiments are conducted based on the following settings. The critics adopt \textbf{4-head} attention networks by default. The actors use feed-forward networks with two hidden layers. For both actors and critics, each hidden layer has 32 neurons. Other hyperparameters are as follows: learning rate of actor is 0.001; learning rate of critic is 0.01; learning rate of target network is $\tau=0.001$; replay buffer size is 100000; batch size is 128; discount factor is $\gamma=0.95$. The network architecture is shown in the Appendix.

\subsection{The Packet Routing Environment}
\begin{figure}[!htb]
    \centering
    \subfigure[The small topology.]{
        \label{fig:twoIE}
        \includegraphics[height=2.6cm]{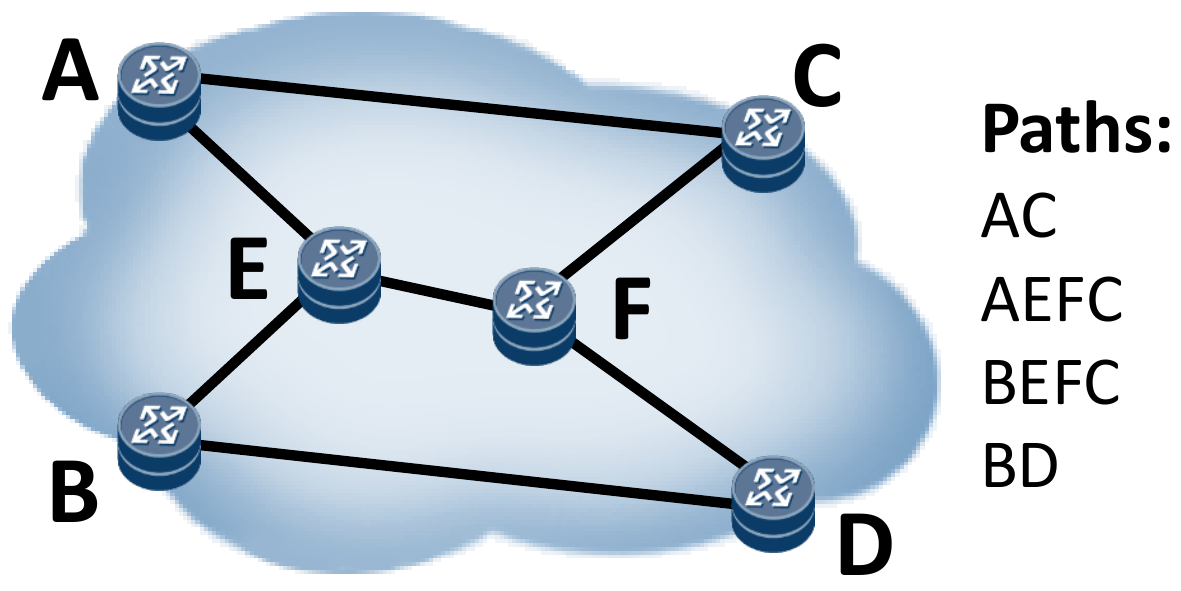}
    }
    \hspace{1.6cm}
    \subfigure[The large topology.]{
        \label{fig:WAN}
        \includegraphics[height=2.6cm]{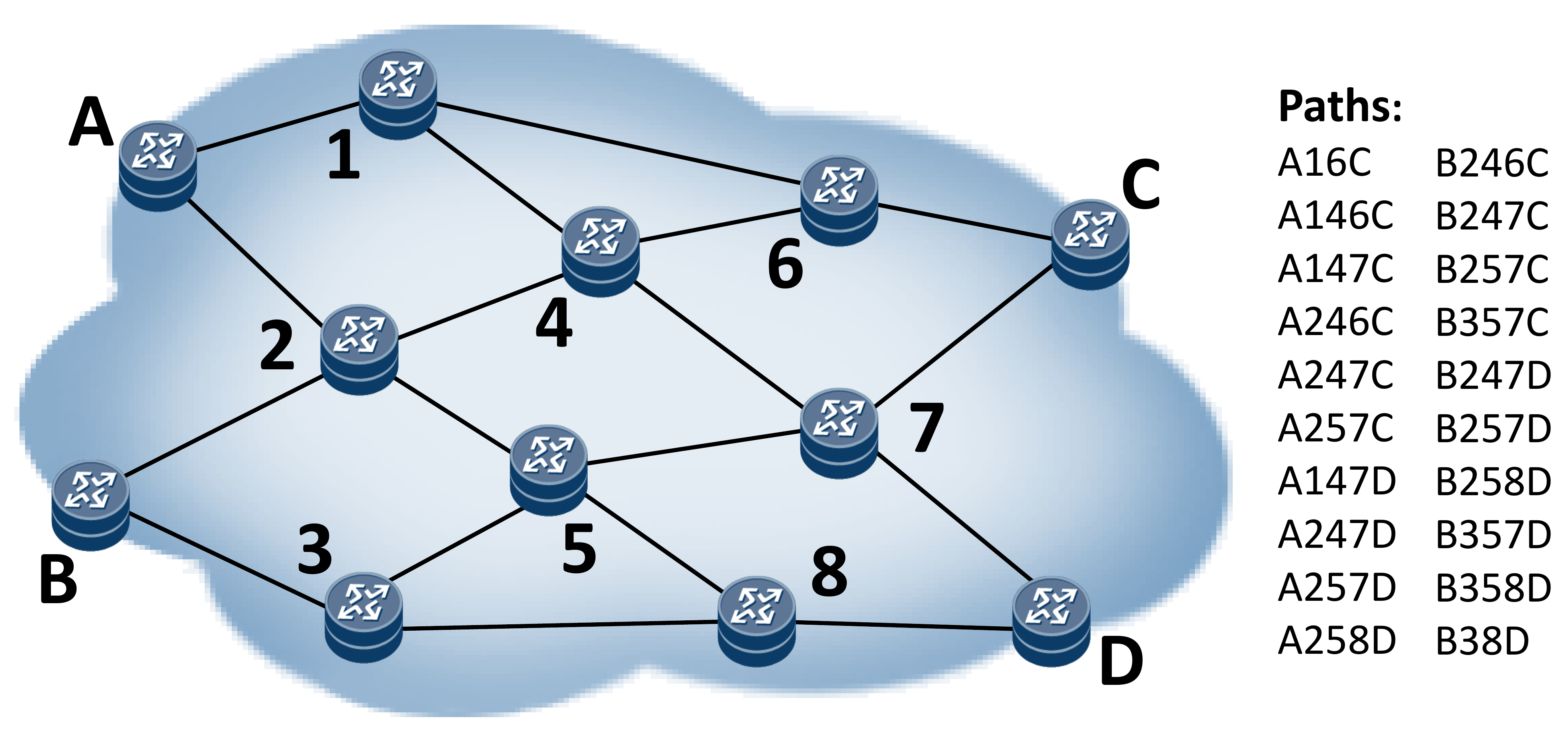}
    }
    \caption{The packet routing environment. The large topology has the same complexity as the real Abilene Network\protect\footnotemark \hspace{-0.06cm} in terms of the numbers of routers, links and paths. It is used for scalability test.}
    \label{fig:PacketRoutingEnvironments}
\end{figure}

\textbf{Environment Description.} Figuring out a better way to route the packets on the Internet is the research topic of our group, so we evaluate our methods on the routing tasks. As shown in Figure \ref{fig:PacketRoutingEnvironments}, the small topology is most classical in the Internet Traffic Engineering community \citep{Kandula2005Walking}; the large topology is the real topology in our application. In each topology, there are several edge routers. Each edge router has an aggregated flow that should be transmitted to other edge routers through available paths (e.g., in Figure \ref{fig:twoIE}, $B$ is set to transmit flow to $D$, and the available paths are $BEFD$ and $BD$). Each path is made up of several links, and each link has a \emph{link utilization}, which equals to the ratio of the current flow on this link to the maximum flow transmission capacity of this link. The necessity of cooperation among routers is as follows: one link can be used to transmit the flow from more than one router, so the routers should not split too much or too little flow to the same link at the same time; otherwise this link will be either overloaded or underloaded.

\footnotetext{A backbone network \url{https://en.wikipedia.org/wiki/Abilene_Network}.}

\textbf{Problem Definition.} The routers are controlled by our algorithm, and they try to learn a good flow splitting policy to minimize the \emph{Maximum Link Utilization in the whole network (MLU)}. The intuition behind this goal is that high link utilization is undesirable for dealing with bursty traffic. The \textbf{\emph{observation}} includes the flow demands in the routers' buffers, the latest ten steps' estimated link utilizations and the latest action taken by the router. The \textbf{\emph{action}} is the splitting ratio of each available path. The \textbf{\emph{reward}} is $1-MLU$ because we want to minimize \emph{MLU}. Exploration bonus based on local link utilization can be added accordingly.

\textbf{Baseline.} MADDPG \citep{lowe2017multi} and PSMADDPGV2 \citep{chu2017parameter} are adopted as baselines, because they are the state-of-the-art RL-based methods that can deal with distributed tasks with continuous action space. They also apply centralized critics to collect teammates' information, but without attention mechanism. MADDPG uses plain fully-connected network to implement the centralized critic, while PSMADDPGV2 uses the parameter sharing method (i.e., sharing part of the critic network with other agents) to model other agents inexplicitly. In addition, the rule-based WCMP and Khead-MADDPG are compared. WCMP \citep{Zhou2014WCMP,Kang2015Efficient} is a Weighted-Cost version of the Equal-Cost Multi-Path routing algorithm\footnote{\url{https://en.wikipedia.org/wiki/Equal-cost_multi-path_routing}.}, which is the most popular multi-path routing algorithm applied in real-world routers. Khead-MADDPG is an ablation model that directly merges the branches of $K$-head Module to generate the real Q-value, and there is no attention mechanism in this model.

\subsubsection{Simple Case Test and Scalability Test.}
\begin{figure}[!htb]
    \centering
    \includegraphics[height=4.6cm]{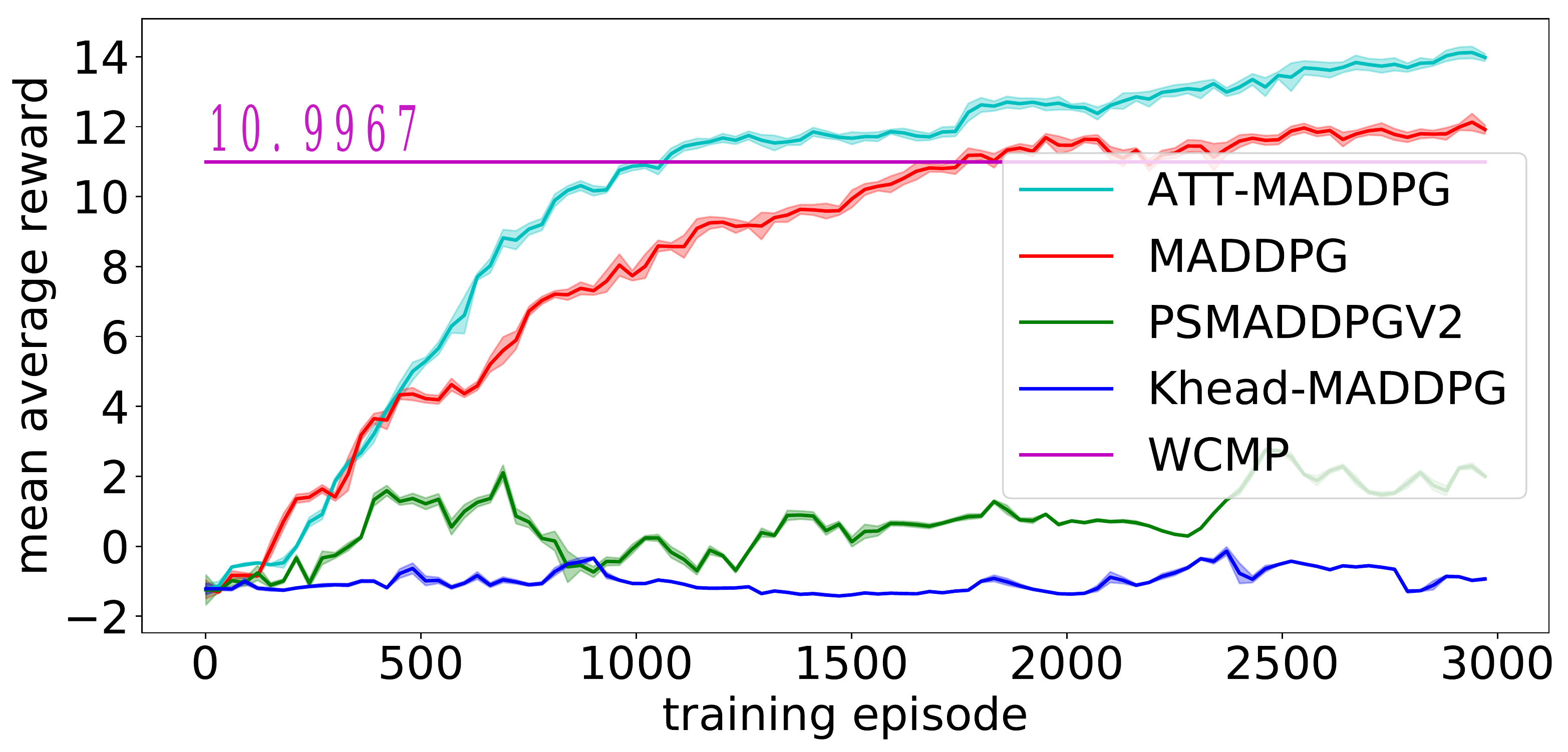}
    \caption{The average rewards on small topology.}
    \label{fig:twoIERewards}
\end{figure}

\begin{figure}[!htb]
    \centering
    \includegraphics[height=4.6cm]{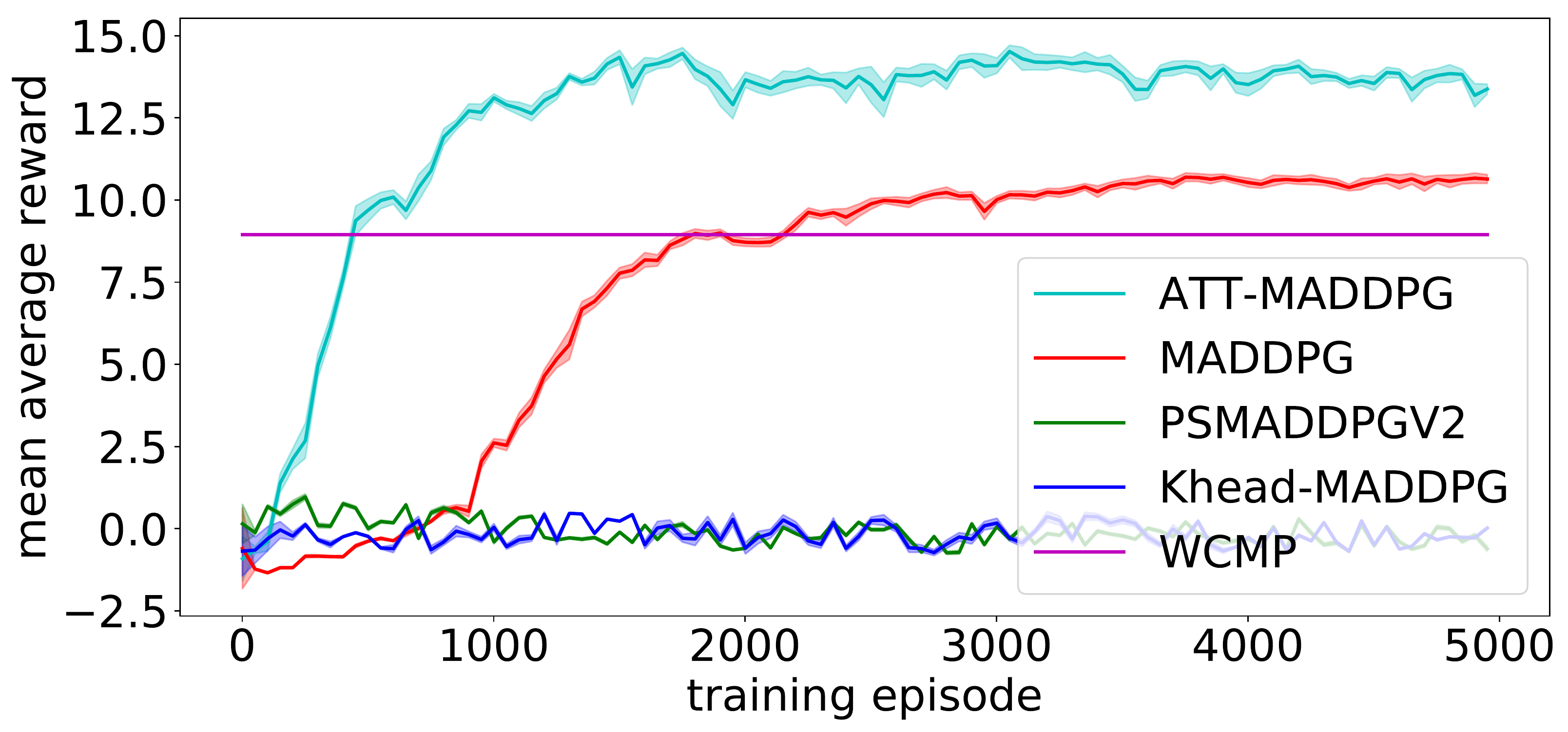}
    \caption{The average rewards on large topology.}
    \label{fig:WANRewards}
\end{figure}

The average rewards of 20 independent experiments are shown in Figure \ref{fig:twoIERewards} and \ref{fig:WANRewards}. As can be seen, for the small topology, ATT-MADDPG can obtain more rewards than MADDPG and PSMADDPGV2, while the Khead-MADDPG model does not work at all. It means that the combination of $K$-head Module and Attention Module (but not a single $K$-head Module) is necessary for achieving good results. The performance of PSMADDPGV2 turns out to be unsatisfactory, which may result from the heterogeneity of the agents.

For the large topology, ATT-MADDPG outperforms other methods by a larger margin. It indicates that ATT-MADDPG has better scalability. A possible reason is that the Attention Module can make the Q-value estimation attend to the actions of more relevant agents (and accordingly, the influence of irrelevant agents is weakened). Take Figure \ref{fig:WAN} as an example, agent4 is very likely to attend to agent1 and agent2 rather than agent3. This property enables ATT-MADDPG to work well even within a complex environment with an increasing number of agents. In contrast, without a mechanism to explicitly model the agents, MADDPG will not be furnished with such scalability.

For both topologies, ATT-MADDPG exhibits better performance than the rule-based WCMP after training a thousand episodes. The reason lies in that the RL-based ATT-MADDPG can take the future effect of actions into consideration, which is in favor of accomplishing the cooperation at a high level, whereas the rule-based WCMP can only consider the current effect of actions.

\subsubsection{Robustness Test.}
\begin{figure}[!htb]
    \centering
    \includegraphics[height=4.6cm]{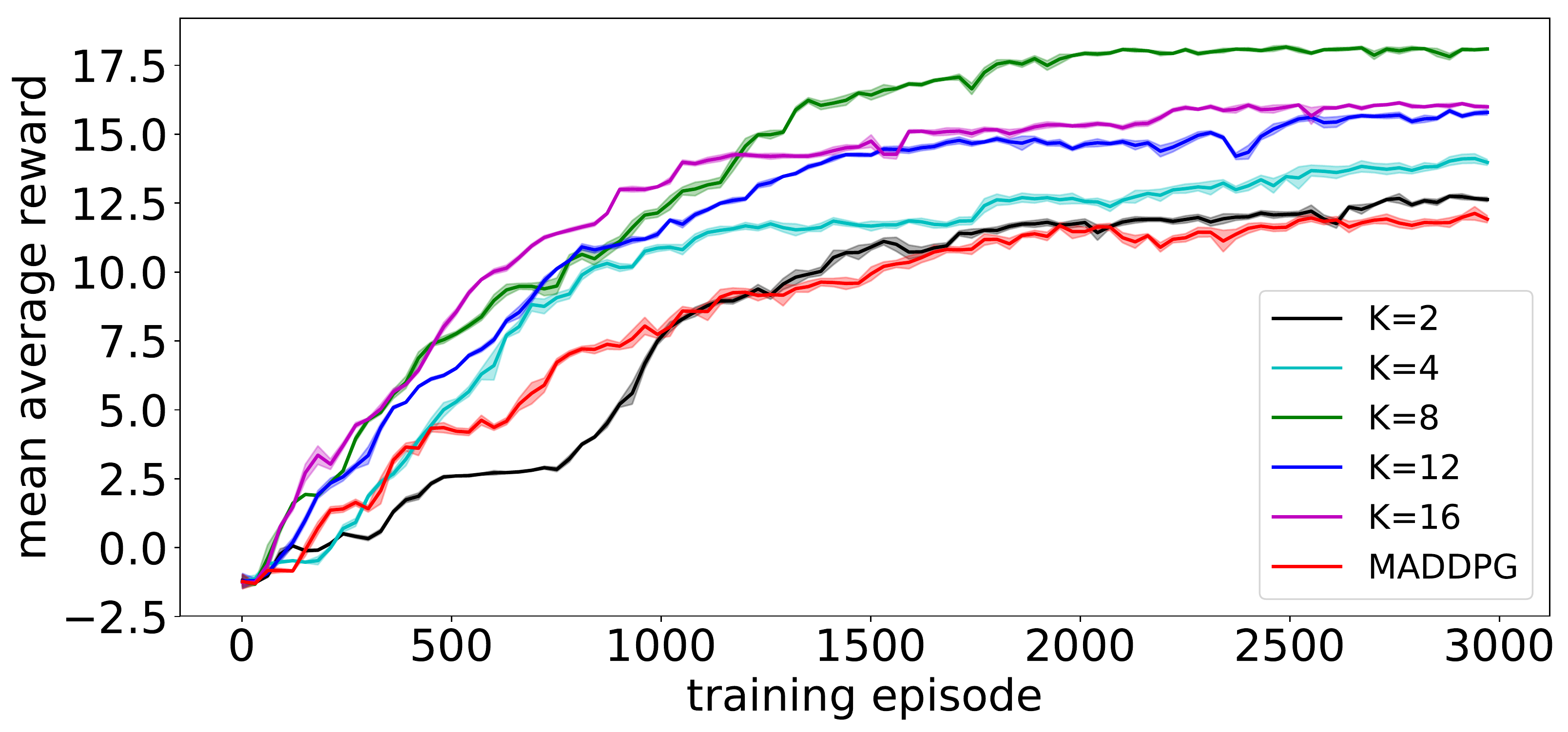}
    \caption{The robustness test on small topology.}
    \label{fig:twoIERewardsDifferentHeads}
\end{figure}

\begin{figure}[!htb]
    \centering
    \includegraphics[height=4.6cm]{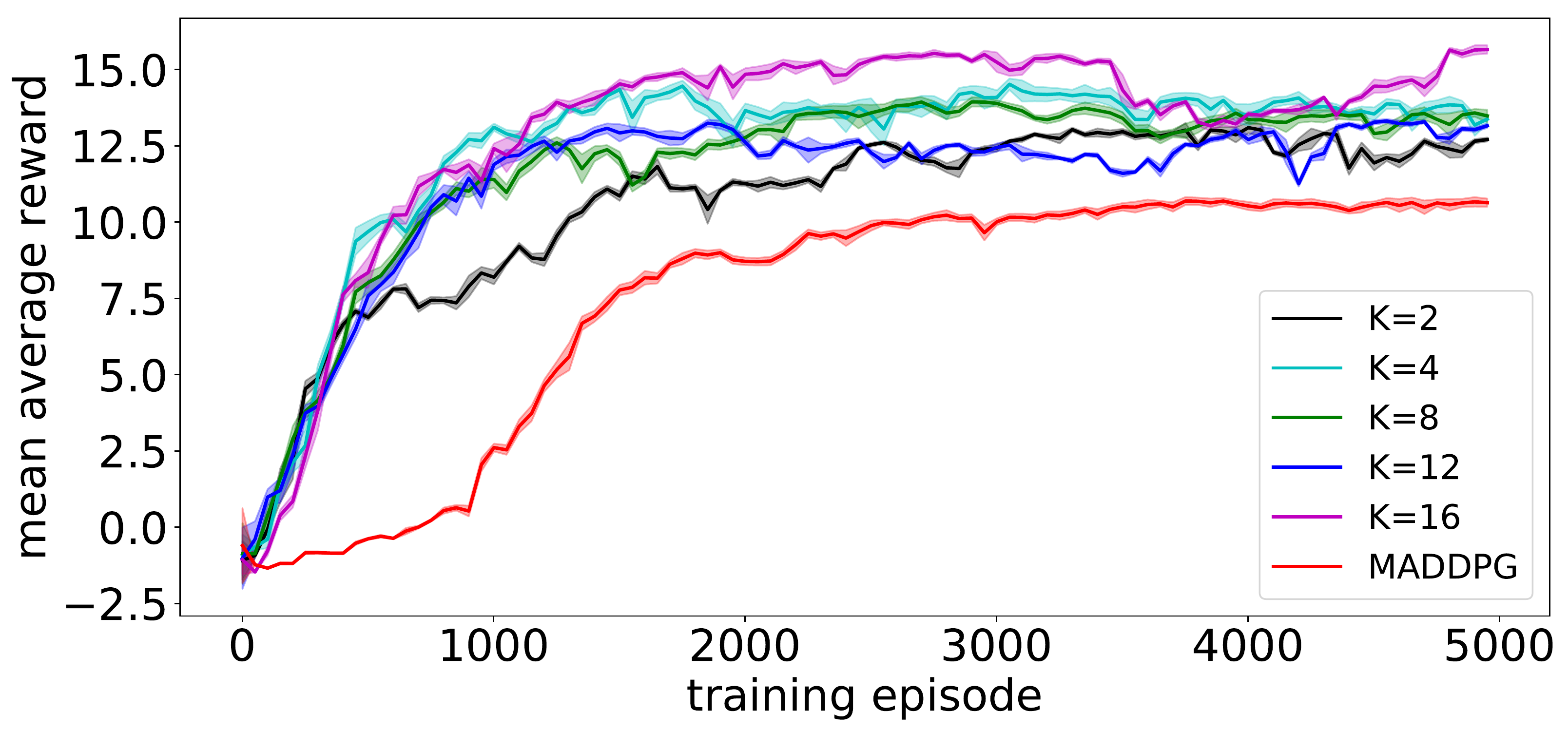}
    \caption{The robustness test on large topology.}
    \label{fig:WANRewardsDifferentHeads}
\end{figure}

ATT-MADDPG introduces a special hyperparameter $K$. It is indispensable to investigate how the setting of $K$ influences the performance. As mentioned before, the above results are obtained when $K=4$. We further set $K$ as 2, 8, 12 and 16 to conduct the same experiments. The average rewards of 20 independent experiments are shown in Figure \ref{fig:twoIERewardsDifferentHeads} and \ref{fig:WANRewardsDifferentHeads}. As can be observed, for the small topology, the obtained rewards are increasing as $K$ becomes greater, and there is a great increase when $K$ is set to 8. For the large topology, a small increase is observed when $K$ is set to 16. Overall, ATT-MADDPG can obtain more rewards than MADDPG in all settings. Consequently, it can be concluded that ATT-MADDPG can stay robust at a wide range of $K$ to achieve good results.

\subsubsection{Further Study on $K$-head and Attention.}\label{AttentionAnalysis}
\begin{figure}[!htb]
    \centering
    \subfigure[The different heads' Q-values.]{
        \label{fig:twoIEQvalueDistribution}
        \includegraphics[height=4.6cm]{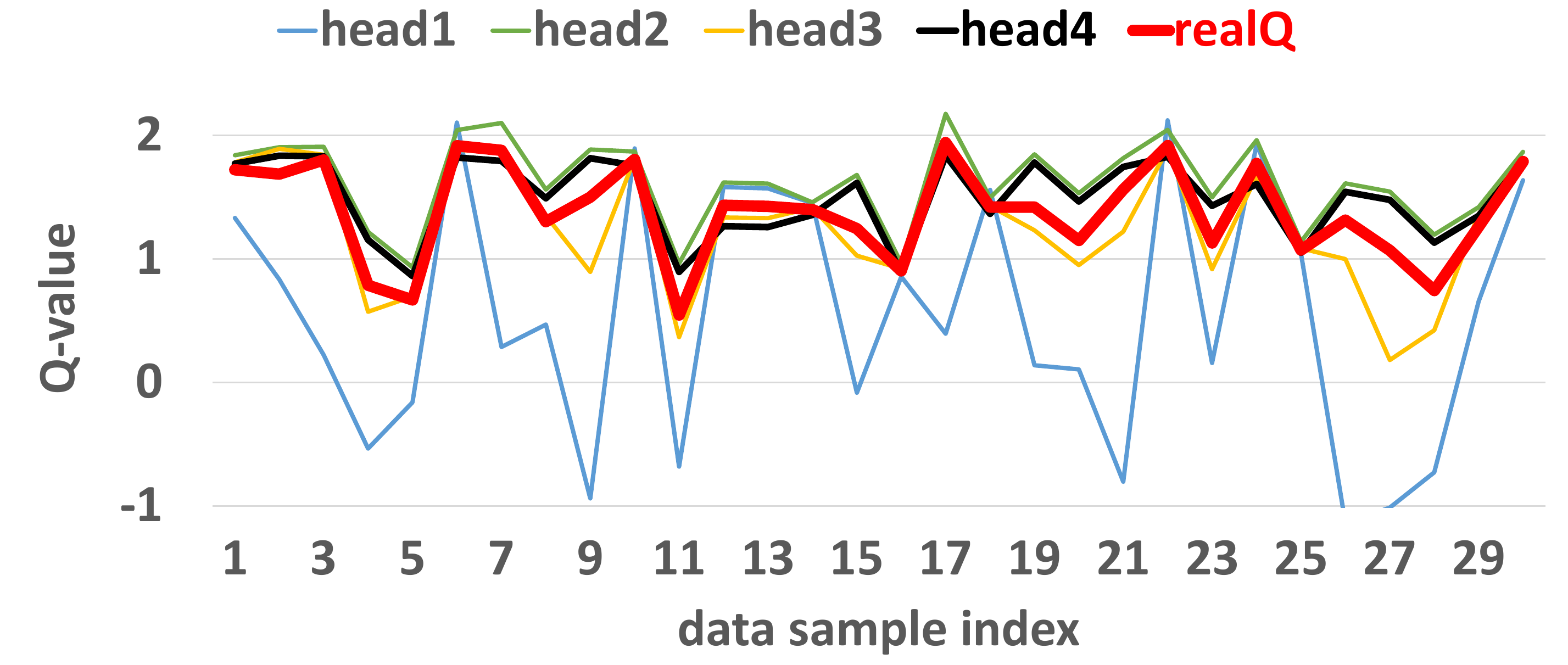}
    }
    \hspace{1.0cm}
    \subfigure[The attention weights.]{
        \label{fig:twoIEAttentionWeight}
        \includegraphics[height=4.6cm]{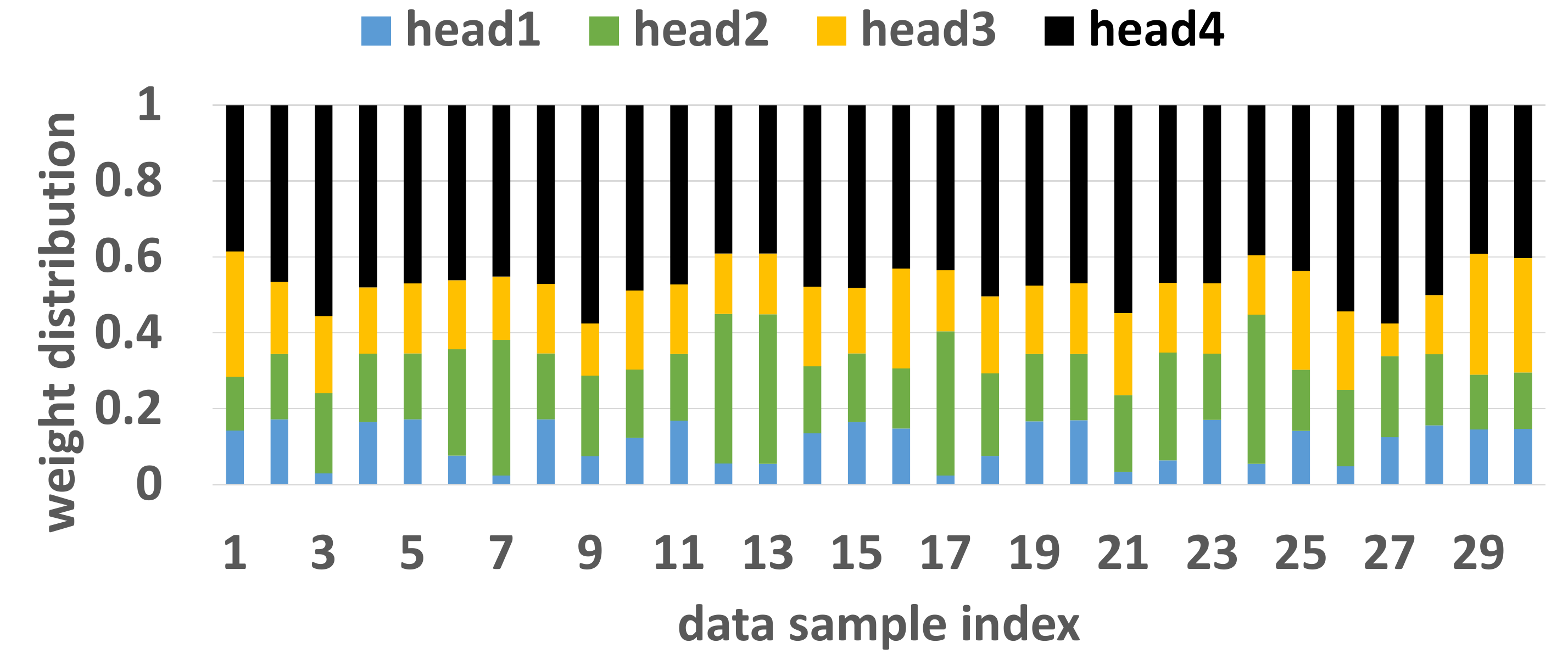}
    }
    \caption{The Q-values and attention weights generated by router $B$ in the small topology.}
    \label{fig:twoIEQvalueDistributionAttentionWeight}
\end{figure}

In Section \ref{Implementation}, we claim that the attention weight $W_i^{k}(w_i)$ is used to approximate the probability $\vec{\pi}_{\bm{-i}}(\vec{a}_{\bm{-i}}|s)$, and the $K$-head Module is expected to have the ability to group similar $\vec{a}_{\bm{-i}}$. In this experiment, we want to verify whether the above claim is consistent with the experimental results. Specifically, we randomly sample 3000 experience tuples $(s,a,Q(s,a))$ from the replay buffer, and show the different heads' Q-values and the attention weights of 30 non-cherry-picked samples\footnote{We only show 30 samples to make the illustration easy to read. To better illustrate all of the 3000 samples, we generate the 2D PCA projections of $(s,a)$, and show a group of 3D figures about $(s,a,Q(s,a))$ in the Appendix. In addition, the Q-value heads are 32D vectors, so we merge the last two layers of the critic network to transform the vector into a scalar Q-value shown in Figure \ref{fig:twoIEQvalueDistribution}. See the Appendix for detailed information.} in Figure \ref{fig:twoIEQvalueDistributionAttentionWeight}. As can be seen, head4 has the smoothest Q-values, and the weights of head4 are much greater than the weights of other heads. In contrast, head1 has a large range of Q-value volatility, and the weights of head1 are much smaller.

The above phenomenon leads us to believe that the $K$-head Module can group similar $\vec{a}_{\bm{-i}}$ indeed. For example, the heavily weighted head4 may represent a large set of non-crucial $\vec{a}_{\bm{-i}}$ (e.g., a flow splitting ratio between [0.3, 0.7]), while the lightly weighted head1 may represent a small set of crucial $\vec{a}_{\bm{-i}}$ (e.g., a flow splitting ratio between [0.8, 0.9]). The explanation is as follows. From the perspective of Q-value, since head4 may represent the \emph{non-crucial} $\vec{a}_{\bm{-i}}$, most local actions $a_{\bm{i}}$ will not have a great impact on the $MLU$ (and accordingly, the reward and the Q-value); therefore it is reasonable that head4 has smooth Q-values. From the perspective of attention weight, as head4 may represent \emph{a large set of} non-crucial $\vec{a}_{\bm{-i}}$ that are preferred by \emph{many} routers, the probability summation $\Sigma_{\vec{a}_{\bm{-i}}} \vec{\pi}_{\bm{-i}}(\vec{a}_{\bm{-i}}|s)$ of the $\vec{a}_{\bm{-i}}$ grouped by head4 will be great; given that the attention weight is an approximation of the probability $\vec{\pi}_{\bm{-i}}(\vec{a}_{\bm{-i}}|s)$, it will be reasonable that head4 has greater attention weights than other heads. The Q-values and the attention weights of head1 can be analysed similarly to show that our hypothesis (i.e., the $K$-head Module can group similar $\vec{a}_{\bm{-i}}$) is reasonable.

\subsection{The Benchmark Environment}
\begin{figure}[!htb]
    \centering
    \subfigure[Cooperative Navigation.]{
        \label{fig:CooperativeNavigation}
        \includegraphics[height=2.6cm,width=3.6cm]{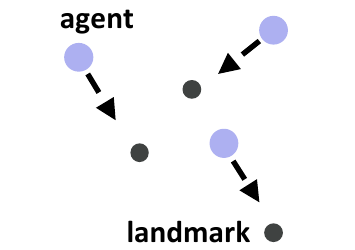}
    }
    \hspace{1.6cm}
    \subfigure[Predator Prey.]{
        \label{fig:PredatorPrey}
        \includegraphics[height=2.6cm]{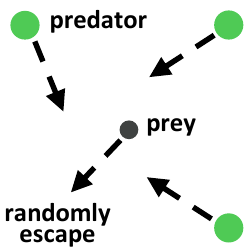}
    }
    \caption{The benchmark environments.}
    \label{fig:BenchmarkEnvironments}
\end{figure}

We consider two benchmark environments that are also adopted by MADDPG. They are shown in Figure \ref{fig:BenchmarkEnvironments}.

\textbf{Cooperative Navigation (Co. Na.).}  Three agents and three landmarks are generated at random locations of a 10-by-10 2D plane. The agents are controlled by our algorithm, and they try to cooperatively cover all landmarks. The \textbf{\emph{observation}} is the relative positions and velocities of other agents and landmarks. The \textbf{\emph{action}} is the velocity. The \textbf{\emph{reward}} is the negative proximity of any agent to each landmark.

\textbf{Predator Prey (Pr. Pr.).} Three predators and a prey are generated at random locations of a 10-by-10 2D plane. The predators are controlled by our algorithm, and they try to cooperatively catch the prey. The \textbf{\emph{observation}} and \textbf{\emph{action}} are the same as those of the cooperative navigation environment. The \textbf{\emph{reward}} is the negative proximity of any predator to the prey. In addition, the predators will get a 10 reward when they catch the prey.

\textbf{Baseline.} Besides MADDPG, PSMADDPGV2 and Khead-MADDPG, we also compare with a rule-based method called GreedyPursuit: for cooperative navigation, the agent always goes to the nearest landmark; for predator prey, the predator always goes to the current location of the prey.

\begin{table}[!htb]
    \centering
    \caption{The average final stable rewards.}
    \label{tab:results-MADDPG-environment}
    \begin{tabular}{|l|c|c|}
        \hline
        & \bf Co. Na. & \bf Pr. Pr. \\
        \hline
        \bf ATT-MADDPG, $\bm{K}$=2 & -1.279 & \bf 3.986 \\
        \hline
        \bf ATT-MADDPG, $\bm{K}$=4 & \bf -1.268 & 3.589 \\
        \hline
        \bf ATT-MADDPG, $\bm{K}$=8 & -1.322 & 3.012 \\
        \hline
        \bf ATT-MADDPG, $\bm{K}$=12 & -1.353 & 3.170 \\
        \hline
        \bf ATT-MADDPG, $\bm{K}$=16 & -1.317 & 3.004 \\
        \hline
        \bf PSMADDPGV2 & -1.586 & 2.473 \\
        \hline
        \bf MADDPG & -1.767 & 1.920 \\
        \hline
        \bf GreedyPursuit & -2.105 & 1.903 \\
        \hline
        \bf Khead-MADDPG & -2.825 & 1.899 \\
        \hline
    \end{tabular}
\end{table}

\begin{figure*}[!htb]
    \centering
    \includegraphics[height=2.8cm]{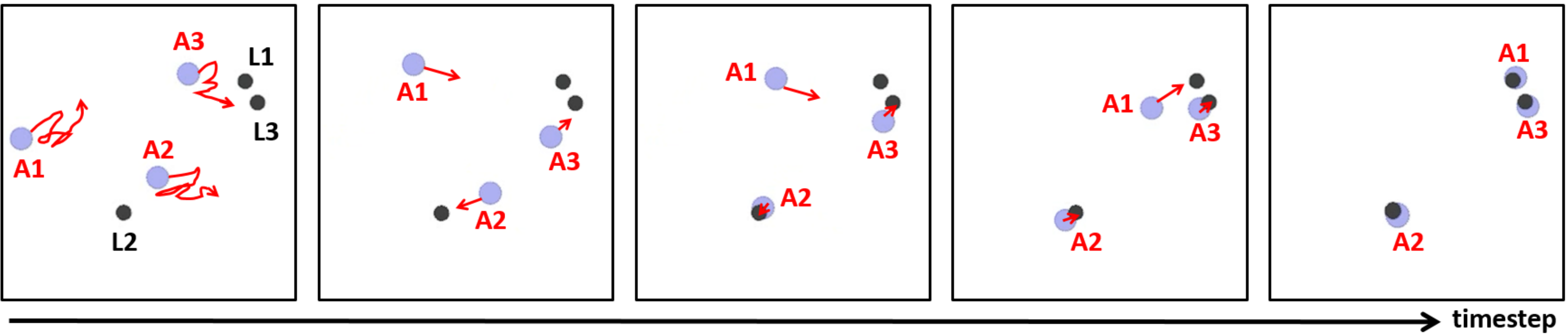}
    \caption{A convergent joint policy learned by ATT-MADDPG under an instance of the cooperative navigation task. L1, L2 and L3 represent different landmarks. A1, A2 and A3 stand for different agents. The red arrows indicate the agents' actions. Note that one picture stands for several timesteps.}
    \label{fig:CooperativeNavigation_PolicyAnalysis}
\end{figure*}

\textbf{The Result.} The average final stable rewards of 50 independent experiments are shown in Table \ref{tab:results-MADDPG-environment}. In contrast to the results in the packet routing environments, PSMADDPGV2 works better than MADDPG in the current environments. The reason may be that the agents are homogeneous in current environments, which makes the parameter sharing method more efficient. Furthermore, ATT-MADDPG can obtain more rewards than MADDPG and PSMADDPGV2 in both environments. It indicates that our method asserts itself with general applicability and good performance. The GreedyPursuit performs badly because it does not consider that the teammates will go to the same landmark, and that the prey will randomly escape to other place. The Khead-MADDPG behaves even worse, because it sometimes cannot converge well, resulting random agents.

\textbf{Policy Analysis.} Figure \ref{fig:CooperativeNavigation_PolicyAnalysis} shows a convergent joint policy learned by ATT-MADDPG under the cooperative navigation task. In the beginning (i.e., the first picture), A1 and A2 share the closest landmark L2, while A3 is very closed to L1 and L3. Therefore, A1 hesitantly moves to the center of L1 and L2, A2 to the center of L2 and L3, A3 to the center of L1 and L3. After some timesteps, the state changes to the second picture. At this point, A2 and A3 understand that A1 will go to L1. Thus, A2 directly moves to L2, A3 to L3, and A1 to L1 in the following timesteps (i.e., the three pictures in the middle). Consequently, the agents cover to all landmarks as shown in the last picture. These behaviors indicate that the agents really learned a cooperative joint policy.

\section{Related Work}
Agent modelling is the process of constructing models for other agents based on the interaction history. The models include any property of interest such as belief, policy, action, class, goal \citep{albrecht2018autonomous}. Most previous methods are based on the Game Theory \citep{ganzfried2011game} or grid-world settings, which are hardly scaled to real-world applications like the network packet routing.

Recently, DRL-based methods has been explored to do agent modelling for large scale problems. Our method is an instance of such method, and the most relevant researches are DRON \citep{cite_DRON}, DPIQN \citep{cite_DPIQN}, LOLA \citep{cite_LOLA}, SOM \citep{cite_SOM}, Mean Field Reinforcement Learning (MFRL) \citep{cite_MeanField}. DRON embeds the opponent's action into the agent's policy network. In this way, the opponent's action can be seen as a hidden variable of the agent's policy. Another gating network is used to control how much the hidden variable influences the policy. DPIQN is very similar to DRON. It embeds the collaborator's policy feature into the controllable agent's DQN \citep{DQN}, such that it is able to generate cooperative actions. LOLA explicitly includes an additional term into the agent's policy updating rules. This additional term can account for the impact to other agents. SOM trains a shared policy network for all agents. The input of the policy network contains a goal field to distinguish different agents. The authors find that the policy network can model the agent's action to some extent. MFRL approximately models the interaction among multiple agents by that between a single agent and the mean effect of other teammates. In contrast to these DQN-based methods that train centralized policies for tasks with discrete action space, our method can generate decentralized policies for tasks with continuous action space. A few DQN-based methods \citep{VDN,QMIX} can generate decentralized policies; the baseline MADDPG \citep{lowe2017multi} and PSMADDPGV2 \citep{chu2017parameter} can train decentralized policies with continuous action space; however, they do not efficiently build models for other agents, instead they address other problems such as credit assignment, competitive agents, and etc. More related studies are shown in the Appendix.

\section{Conclusion}
This paper presents a novel actor-critic RL method to model and exploit teammates' policies in the cooperative distributed multi-agent setting. Our method embeds an attention mechanism into a centralized critic, which introduces a special structure to \emph{explicitly} model the dynamic \emph{joint policy} of teammates in an \emph{adaptive manner}. Consequently, all agents will cooperate with each other efficiently. Furthermore, our method can train \emph{decentralized policies} to handle distributed tasks with \emph{continuous action space}.

We evaluate our method on both benchmark tasks and the real-world packet routing tasks. The results show that it not only outperforms the state-of-the-art RL-based methods and rule-based methods by a large margin, but also achieves good scalability and robustness. Moreover, to better understand our method, we also make thorough experiments: (1) the ablation model illustrates that all components of the proposed model are necessary; (2) the study on Q-values and attention weights demonstrates that our method has mastered a sophisticated attention mechanism indeed; (3) the analysis of a concrete policy shows that the agents really learned a cooperative joint policy.

Future work will extend our method to the settings with discrete action space and competitive agents.

\subsubsection*{Acknowledgments}
The authors would like to thank prof. Zhihua Zhang for helpful suggestions. The authors would also like to thank Yan Ni, Shiru Ren, Xiangyu Liu, Yuanxing Zhang, Shihan Xiao and the anonymous reviewers for their insightful comments. This work was supported by the National Natural Science Foundation of China under Grant No.61572044. The contact author is Zhen Xiao.

\bibliographystyle{bibliographystyle-ACM}  
\bibliography{bibliography-reference}  

\end{document}